\documentclass[letterpaper, 11 pt]{article} 
\usepackage{amsmath,amsfonts,amssymb,color,amsthm}
\usepackage{nccmath}
\usepackage{epsfig}
\usepackage{psfrag}
\usepackage[T1]{fontenc}
\usepackage{algorithm}
\usepackage{algpseudocode}
\usepackage[dvipsnames]{xcolor}
\usepackage{array}
\usepackage{epstopdf}
\usepackage{cite}
\usepackage{footmisc}
\usepackage{mathtools}
\usepackage{bbm}
\usepackage{bm}
\usepackage{comment}
\usepackage{dsfont}

\usepackage{mathtools}
\usepackage{epstopdf}
\usepackage{tikz}
\usetikzlibrary{automata, positioning, arrows.meta}
\usepackage{relsize}
\usetikzlibrary{shapes,arrows}
\usepackage{cite}
\usepackage{epstopdf}
\usepackage{tcolorbox}
\definecolor{winered}{rgb}{0.5,0,0}
\usepackage{scalerel}
\usepackage{xcolor}
\usepackage[colorlinks]{hyperref}
\usepackage{tcolorbox}
\tcbset{colback=blue!5!white,colframe=blue}
\newcommand\R{\bar{r}}
\newcommand{\mc}{\mathcal}

\AtBeginDocument{%
  \hypersetup{
    citecolor=blue,
    linkcolor=winered
  }
}

\DeclarePairedDelimiter\ceil{\lceil}{\rceil}

\newtheorem{problem}{Problem}

\newtheorem{theorem}{Theorem}

\newtheorem{lemma}{Lemma}
\newtheorem{remark}{Remark}

\date{}
\usepackage{cite}
\usepackage[margin=1in]{geometry}

\title{\LARGE \bf Robust Q-Learning under Corrupted Rewards}

\author{Sreejeet Maity and Aritra Mitra} % Use a specific footnote mark

\begin{document}
\maketitle
\footnotetext[1]{The authors are with the Department of Electrical and Computer Engineering, North Carolina State University. Email: {\tt \{smaity2, amitra2\}@ncsu.edu}.}
\begin{abstract} Recently, there has been a surge of interest in analyzing the non-asymptotic behavior of model-free reinforcement learning algorithms. However, the performance of such algorithms in non-ideal environments, such as in the presence of corrupted rewards,  is poorly understood. Motivated by this gap, we investigate the robustness of the celebrated Q-learning algorithm to a strong-contamination attack model, where an adversary can arbitrarily perturb a small fraction of the observed rewards. We start by proving that such an attack can cause the vanilla Q-learning algorithm to incur arbitrarily large errors. We then develop a novel robust synchronous Q-learning algorithm that uses historical reward data to construct robust empirical Bellman operators at each time step. Finally, we prove a finite-time convergence rate for our algorithm that matches known state-of-the-art bounds (in the absence of attacks) up to a small inevitable $O(\varepsilon)$ error term that scales with the adversarial corruption fraction $\varepsilon$. {Notably, our results continue to hold even when the true reward distributions have infinite support, provided they admit bounded second moments.}

\end{abstract}
\section{Introduction}
We study reinforcement learning (RL) within a Markov decision process (MDP) setting where an agent sequentially interacts with an environment to maximize a long-term cumulative value function. To achieve this goal without knowledge of the dynamics of the MDP, the agent plays an action at each time, receives a reward in the form of feedback from the environment, and then transitions to a new state. This process then repeats itself. Using the sequence of observed rewards, the agent learns to take ``better" actions over time. In this context, one of the most widely studied model-free RL algorithms is the celebrated Q-learning algorithm of Watkins and Dayan~\cite{watkins1992q} that  iteratively estimates the optimal state-action value function associated with the MDP. Viewing Q-learning through the lens of stochastic approximation (SA) theory, a rich body of work has investigated the asymptotic performance of this algorithm, i.e., its behavior as the number of iterations goes to infinity~\cite{borkar, tsitsiklis94, jaakkola, csabaasymp}. More recently, there has been a shift of interest towards providing a finer non-asymptotic/finite-time analysis of different RL algorithms~\cite{bhandari2018finite, shah2018, srikantfinite, Waiwright, Qu, li2024q}. That said, the literature discussed above exclusively pertains to scenarios where the reward feedback received from the environment is \emph{always accurate}, i.e., it is generated by the true reward distribution of the underlying MDP. However, such an idealistic assumption is unlikely to hold when one seeks to deploy RL algorithms in real-world harsh environments. As such, since feedback in the form of rewards is crucial for the overall  decision-making process, our main goal in this paper is to investigate the following questions: 
\begin{tcolorbox}
    \emph{Q1. Are existing model-free RL algorithms robust to perturbations in the rewards?}\\
   \emph{Q2. Can we obtain reliable value-function estimates under corrupted rewards?}
\end{tcolorbox}  

\textbf{Contributions.} In response to the aforementioned questions, the main contributions of this work are as follows. 

$\bullet$ \textbf{Problem Formulation.} To systematically investigate the question of robustness of RL algorithms, we consider a \emph{strong-contamination reward attack model} where an adversary with complete knowledge of the MDP and the learner's observations is allowed to \emph{arbitrarily} perturb $\varepsilon$-fraction of the rewards acquired over time. This attack model is directly inspired from the robust statistics literature~\cite{huber, huber2, lugosi}, where a small fraction of the data points in a static data set can be corrupted by an adversary. 

$\bullet$ \textbf{Vulnerability of vanilla Q-Learning.} In Theorem~\ref{thm:vulnerability}, we prove that even under a weaker attack model than the one described above, the basic Q-learning algorithm converges to the fixed point of a perturbed Bellman operator, where the perturbation is due to reward contamination. By constructing an explicit example, we next establish in Theorem~\ref{thm:example} that this incorrect fixed point can be arbitrarily far away from the optimal Q function. Furthermore, this continues to be the case even when the corruption fraction $\varepsilon$ is small. 

$\bullet$ \textbf{Robust Q-Learning Algorithm.} Motivated by the above findings, we develop a novel robust Q-learning algorithm in the synchronous sampling setting~\cite{kearns}, i.e., when a generative model provides independent samples for all state-action pairs. The key idea in our algorithm is to use historical data of rewards for each state-action pair to construct a robust empirical Bellman operator in each iteration. To do so, we leverage the recently developed robust trimmed mean-estimator in~\cite{lugosi}, along with a novel dynamic thresholding technique.  

$\bullet$ \textbf{Near-optimal Rates.} Providing a rigorous finite-time convergence analysis of our algorithm is non-trivial since one needs to simultaneously contend with the benign randomness introduced by sampling and deliberate arbitrary adversarial injections. Nonetheless, in Theorem~\ref{theorem:theoremmainresult}, we establish a high-probability $\ell_\infty$-error-rate of $\tilde{O}(1/\sqrt{T})+O(\sqrt{\varepsilon}),$ where $T$ is the number of iterations. In the absence of corruption (i.e., $\varepsilon=0$), this rate matches those in recent works~\cite{Waiwright, Qu}. Furthermore, we also provide an informal argument that the additive $O(\sqrt{\varepsilon})$ term appears to be inevitable.

{Finally, we note that our proposed approach does not require the true reward distributions to be bounded or ``light-tailed", i.e., we do not need to make any assumptions of sub-Gaussianity on the reward models. Instead, we only need the reward distributions to admit finite second moments. Thus, in principle, even the true reward samples (i.e., the inliers) can come from heavy-tailed distributions. This makes it non-trivial to tell apart clean reward data from corrupted rewards. Nonetheless, the approach developed in this paper overcomes this challenge.}

\textbf{Related Work.} The synchronous Q-learning setting we consider here has been extensively studied in several works~\cite{kearns, even2003learning, sidford}. More recently, its finite-time performance has been characterized in~\cite{shah2018, Waiwright, li2024q}. Asynchronous versions of Q-learning have also been analyzed  in~\cite{beck2012, Qu, li2024q}. 

Reward corruption models similar to what we consider here have been studied in the multi-armed bandits (MAB) literature~~\cite{junadv, lykouris, liuadv, guptaadv, kapoor, bogunovic1, bogunovic2}, where an attacker with a finite budget can corrupt the rewards seen by the learner in a small fraction of the rounds. However, our setup, algorithm, and proof technique fundamentally differ from the MAB framework. That said, our main result in Theorem~\ref{theorem:theoremmainresult} has the same flavor as the findings in the above bandit papers: one can recover the performance bounds in the absence of corruptions up to an additive term that captures the attacker's budget; in our case, this is the additive $O(\sqrt{\varepsilon})$ term. Finally, some recent works have studied perturbed versions of Markovian stochastic approximation algorithms~\cite{mitraEF, adibi2024}, where the perturbations are due to communication constraints (e.g., delays and compression). Unlike the \emph{structured} perturbations in these papers, the perturbations injected by the adversary in our work can be \emph{arbitrary}.  

\section{Background and Problem Formulation}
\label{sec:prob_form} 
\textbf{MDP Model.} We consider a Markov Decision Process (MDP) denoted by $\mathcal{M}=(\mathcal{S}, \mathcal{A}, {P}, R, \gamma)$, where $\mathcal{S}$ is a finite state space, $\mathcal{A}$ is a finite action space, ${P}$ is a set of Markov transition kernels, $R$ is a reward function, and $\gamma \in (0,1)$ is the discount factor. Upon playing action $a$ at state $s$, the state of the MDP transitions to $s'$ with probability ${P}(s'|s,a)$, and a scalar stochastic immediate reward $r(s,a)$ with distribution $\mc{R}(s,a)$ is observed. We define $R(s,a) \triangleq \mathbb{E}[r(s,a)]$ to be the expected value of the random variable $r(s,a)$, and assume that the (noisy) rewards are bounded, i.e., $\exists \bar{r} \geq 1$ such that $|r(s,a)| \leq \bar{r}, \forall (s,a) \in \mc{S} \times \mc{A}$.\footnote{{The assumption of bounded rewards is made here only to convey the key ideas in their simplest form. Later, in Section~\ref{sec:unboundedrewards}, we will see how our techniques naturally allow for reward distributions with finite second moments that can have potentially unbounded support.}} We consider deterministic policies $\pi: \mathcal{S} \rightarrow \mathcal{A}$ that map states to actions. The ``goodness" of a policy $\pi$ is captured by a $\gamma$-discounted infinite-horizon cumulative reward $V_{\pi}: \mc{S} \mapsto \mathbb{R}$ given by:
\begin{equation}\label{eqn:V_reward}
V_{\pi}(s) = \mathbb{E}\left[\sum_{t=0}^{\infty} \gamma^t r(s_t,a_t) \, \Big| \, s_0 = s\right],
\end{equation}
where $s_t$ is the state at time $t$, $a_t = \pi(s_t)$ is the action played at time $t$, and the expectation is taken w.r.t. the randomness in the states and rewards. We will refer to $V_\pi(s)$ as the value function corresponding to state $s$. Given this premise, the goal of the learner is to find a policy $\pi$ that maximizes $V_\pi(s)$ simultaneously for all states $s \in \mc{S}.$ It is well known that such a deterministic optimal policy $\pi^*$ does exist~\cite{suttonRL}. Let us now briefly discuss how such a policy can be obtained when the MDP is known. To that end, we define the state-action value function $Q_{\pi}:\mc{S} \times \mc{A} \mapsto \mathbb{R}$ as follows: 
\begin{equation}\label{eqn:Q_reward}
Q_{\pi}(s, a) = \mathbb{E}\left[\sum_{t=0}^{\infty} \gamma^t r(s_t,a_t) \, \Big| \, (s_0, a_0) = (s, a)\right].
\end{equation}
Now let $Q^{*} = Q_{\pi^*}$ denote the optimal state-action value function. Then, $Q^*$ is the unique fixed point of the Bellman operator $\mathcal{T}^{*}: \mathbb{R}^{|\mc{S}|  \times |\mc{A}|}  \rightarrow \mathbb{R}^{|\mc{S}|  \times |\mc{A}|}$ given by:
\begin{equation}\label{eqn:Bellman}
    (\mathcal{T}^{*}Q)(s,a) = R(s,a) + \gamma \mathbb{E}_{s' \sim {P}(\cdot | s,a)}\left[\max_{a' \in \mathcal{A}} Q(s',a')\right].
\end{equation}
In other words, $\mc{T}^* (Q^*) = Q^*.$ The Bellman operator satisfies the following contraction property $\forall Q_1, Q_2 \in \mathbb{R}^{|\mc{S}| \times |\mc{A}|}$:
\begin{equation}\label{eqn:Bellmancontraction}
\lVert \mathcal{T}^* (Q_1) - \mathcal{T}^* (Q_2)\rVert_{\infty}\le \gamma \lVert Q_1 - Q_2\rVert_{\infty}.
\end{equation}

An immediate consequence of the above property is that the iterative update rule $Q_{t+1} = \mc{T}^*(Q_t)$ guarantees exponentially fast convergence to $Q^*$. This is precisely the idea used in dynamic programming~\cite{suttonRL}. In our RL setting, however, the dynamics of the MDP are unknown, rendering the above idea infeasible. We now discuss a synchronous version~\cite{kearns, even2003learning, sidford} of the seminal Q-learning algorithm that finds $Q^*$ \emph{without knowledge} of the MDP dynamics. 

\textbf{Synchronous Q-learning.} The synchronous Q-learning algorithm operates in iterations $t=0, 1, \ldots$, where in each iteration $t$, the learner maintains an estimate $Q_t$ of $Q^*$. The synchronous setting assumes the existence of a generative model such that in each iteration $t$, the learner gets to observe the following objects for each state-action pair $(s,a) \in \mc{S} \times \mc{A}$: (i) a new state $s_t(s,a)$ drawn independently from ${P}(\cdot| s,a)$; and (ii) a stochastic reward $r_t(s,a)$ drawn independently from $\mc{R}(s,a)$. Using this information, for each $(s,a) \in \mc{S} \times \mc{A}$, the learner updates $Q_t(s,a)$ as follows: 
\begin{equation}
\label{eqn:syncQ}
Q_{t+1}(s,a) = (1- \alpha_t)Q_t(s,a) + \alpha_t  (\mc{T}_tQ_t)(s,a), 
\end{equation}
where $\{\alpha_t\}$ is a suitable step-size sequence, and $\mc{T}_t:\mathbb{R}^{|\mc{S}| \times |\mc{A}|} \rightarrow \mathbb{R}^{|\mc{S}| \times |\mc{A}|}$ is an empirical Bellman operator constructed from observations at iteration $t$, and defined as
$$ (\mc{T}_tQ)(s,a) \triangleq r_t(s,a) + \gamma \max_{a'\in \mathcal{A}}Q(s_t(s,a),a'), \forall Q \in \mathbb{R}^{|\mc{S}| \times |\mc{A}|},$$
where $r_t(s,a) \sim \mc{R}(s,a)$, and $s_t(s,a) \sim {P}(\cdot| s,a).$ The term ``synchronous" arises from the fact that in each iteration $t$, the learner gets to observe independent data samples for \emph{every} state-action pair. As such, every component of the vector $Q_t$ can be updated at iteration $t$, i.e., synchronously. Viewing synchronous Q-learning as a stochastic approximation (SA) scheme, one can show that under suitable assumptions on the step-size sequence $\{\alpha_t\}$, $Q_t \rightarrow Q^*$ with probability $1$~\cite{tsitsiklis94, csabaasymp}. Our goal in this paper is to provide a finite-time analysis of synchronous Q-learning when the observed rewards can be potentially corrupted by an omniscient adversary. We formally describe our attack model below. 

\textbf{Strong-Contamination Reward Attack Model.} We consider an adversary that has complete knowledge of the MDP model and the observations of the learner in every iteration. Using this information, in each iteration $t$, the adversary can perturb the entire reward set $\{r_t(s,a)\}_{(s,a) \in \mc{S} \times \mc{A}}$ \emph{arbitrarily}. However, to make the problem meaningful, we will associate an attack budget $\varepsilon \in [0, 1/2)$ with the attacker: for each $t \in \mathbb{N}$, the attacker is allowed to corrupt the reward sets in at most $\varepsilon$-fraction of the first $t$ iterations. Our attack model is directly inspired by the strong contamination model from the robust statistics literature~\cite{huber, huber2, lugosi} where an adversary is allowed to arbitrarily perturb at most $\varepsilon$-fraction of the data points in a set; in our context, the reward observations constitute the data. We note that our model above is more powerful than the classical Huber attack model~\cite{huber, huber2} where each data point can be corrupted with probability $\varepsilon$. In particular, the Huber model would imply that $\varepsilon t$ reward sets are corrupted \emph{only on an average}, for each $t\in \mathbb{N}$. In the sequel, we will use $y_t(s,a)$ to denote the observed reward for state-action pair $(s,a)$ in iteration $t$. In an iteration $t$ where there is no corruption, $y_t(s,a)=r_t(s,a), \forall (s,a) \in \mc{S} \times \mc{A}.$ 
\begin{tcolorbox}
\begin{problem} Given the strong-contamination reward attack model and a failure probability $\delta \in (0,1)$, our goal is to generate a robust estimate ${Q}_t$ of the optimal value function $Q^*$ such that with probability at least $1-\delta$, the $\ell_{\infty}$-error $\Vert {Q}_t - Q^* \Vert_{\infty}$ is bounded from above by an error-function $e(t,\varepsilon)$ that has optimal dependence on the number of iterations $t$ and corruption fraction $\varepsilon$. 
\end{problem}
\end{tcolorbox}
In the next section, we show that the vanilla synchronous Q-learning algorithm fails to achieve this goal. In Section~\ref{sec:Algorithm}, we then proceed to develop our proposed robust algorithm.

\section{Motivation}
In this section, we will show that even with a small attack budget $\varepsilon$, an adversary can cause the vanilla Q-learning algorithm to converge to a point in $\mathbb{R}^{|\mathcal{S}|\times|\mathcal{A}|}$ arbitrarily far away from $Q^*$. To do so, it suffices to consider the Huber attack model~\cite{huber, huber2}. Accordingly, in each iteration $t$, we toss a biased coin with probability of heads $1-\varepsilon$. If the coin lands heads, the observed reward $y_t(s,a)$ is drawn from the true reward distribution $\mc{R}(s,a), \forall (s,a) \in \mc{S} \times \mc{A}.$ If it lands tails, $y_t(s,a)$ is drawn from an arbitrary distribution $\mc{C}(s,a), \forall (s,a) \in \mc{S} \times \mc{A}.$ Concretely, $y_t(s,a) \sim (1-\varepsilon) \mc{R}(s,a) + \varepsilon \mc{C}(s,a).$ Under this Huber attack model, suppose the perturbed expected reward for $(s,a)$ is given by $\tilde{R}_{c}(s,a) = \mathbb{E}[y_t(s,a)].$ We have the following simple result. 
\begin{tcolorbox}
\begin{theorem}\label{thm:vulnerability} Consider the vanilla synchronous Q-learning algorithm in Eq.~\eqref{eqn:syncQ} with rewards perturbed based on the Huber attack model described above. If the step-size sequence satisfies $\alpha_t \in (0,1)$ with  $\sum_{t=1}^{\infty}\alpha_t = \infty$ and $\sum_{t=1}^{\infty} \alpha_t^2 < \infty$, then with probability $1$, $Q_t \rightarrow \tilde{Q}^*_c$, where $\tilde{Q}^*_c$ is the unique fixed point of the perturbed Bellman operator $\mathcal{\tilde{T}}^*_c: \mathbb{R}^{|\mc{S}|  \times |\mc{A}|}  \rightarrow \mathbb{R}^{|\mc{S}|  \times |\mc{A}|}$ defined by 
\begin{equation}\label{eqn:perturb_Bellman}
    (\mathcal{\tilde{T}}^*_c Q)(s,a) = \tilde{R}_c(s,a) + \gamma \mathbb{E}_{s' \sim {P}(\cdot | s,a)}\left[\max_{a' \in \mathcal{A}} Q(s',a')\right].
\end{equation}
\end{theorem}
\end{tcolorbox}
\begin{proof}
It suffices to simply note that under the Huber attack model, we end up running the synchronous Q-learning algorithm on a new MDP $(\mathcal{S}, \mathcal{A}, {P}, \tilde{R}_c, \gamma)$ that differs from the original MDP only in its reward function, where $\tilde{R}_c$ is the perturbed reward function defined earlier. The claim then follows directly by appealing to the asymptotic convergence of synchronous Q-learning based on SA theory~\cite{tsitsiklis94, csabaasymp}. 
\end{proof}

Theorem~\ref{thm:vulnerability} tells us that under the Huber model, an adversary can bias the iterates generated by vanilla Q-learning towards the fixed point $\tilde{Q}^*_c=\mathcal{\tilde{T}}^*_c(\tilde{Q}^*_c)$ of a perturbed Bellman operator $\mathcal{\tilde{T}}^*_c$. However, it does not provide an explicit lower bound on the gap $\Vert \tilde{Q}^{*}_c -Q^* \Vert_{\infty}.$ Our next result reveals that this gap can be arbitrarily large. 

\begin{figure}[t]
\begin{center}
\begin{tikzpicture}[->, >=stealth', auto, thick, node distance=2.8cm]
	\tikzstyle{every state}=[fill=white,draw=black,thick,text=black,scale=1]
	\node[state]    (A)                     {$2$};
	\node[state]    (B)[right of=A]   {$1$};
	\node[state]    (C)[right of=B]   {$3$};
	\node[state]    (D)[above of=C]   {$4$};
    \node[state]    (E)[above of=A]   {$5$};
	\path
	(A) edge[loop left] node{$p$} (A)
    (B) edge[bend left,below] node{$p, \texttt{\textcolor{red}{L}}$} (A)
        edge[bend right,below] node{$p, \texttt{\textcolor{red}{R}}$} (C)
        edge[bend right,left] node{$1-p,\texttt{\textcolor{red}{L}}$} (E)
        edge [bend left, right] node{$1-p,\texttt{\textcolor{red}{R}}$} (D) % New edge from state 1 to state 4
    (C) edge[bend right,right] node{$1-p$} (D)
        edge[loop right] node{$p$} (C)
    (D) edge[loop right] node{$1$} (D)
    (E) edge[loop left] node{$1$} (D)
    (A) edge[bend left,left] node{$1-p$} (E); % New edge from state 1 to state 5
	%\node[above=0.5cm] (A){Patch G};
	%\draw[red] ($(D)+(-1.5,0)$) ellipse (2cm and 3.5cm)node[yshift=3cm]{Patch H};
	\end{tikzpicture}
\caption{The above MDP is constructed with state space $\mathcal{S} =\{1, 2, 3, 4, 5\}$ and action space $\mathcal{A} = \{\texttt{L}, \texttt{R}\}$. When in state $s=1$, taking action $a = L$ leads to a transition to state 2 with probability $p$, and a transition to state 5 with probability $1-p$. Taking action $a = R$ in state $s=1$ leads to symmetric outcomes. In states 2 and 3, regardless of the chosen action, the system remains in states 2 and 3 with probability $p$, and transits to states 4 and 5 with probability $1-p$. States 4 and 5 are absorbing states, indicating that once reached, the system remains in these states indefinitely.} \label{fig:MDP}
\end{center}
\end{figure}
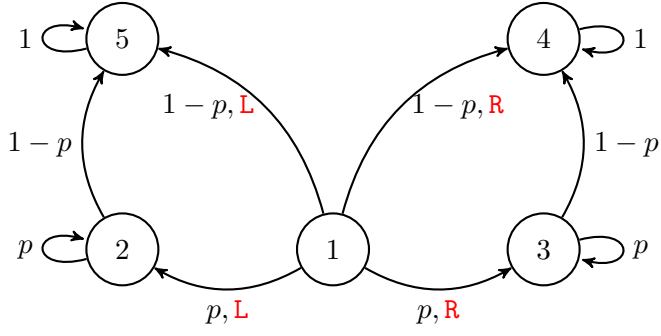
\begin{tcolorbox}
\begin{theorem} 
\label{thm:example}
There exists an MDP with finite state and action spaces for which the gap \\
$\Vert \tilde{Q}^{*}_c -Q^* \Vert_{\infty}$ can be arbitrarily large under the Huber attack model. 
\end{theorem}
\end{tcolorbox}
\begin{proof}
We will prove this result by constructing an explicit example that is inspired from \cite{azar2012sample}. To that end, consider the MDP in Fig.~\ref{fig:MDP}. We first describe the reward model without attacks. In state $s =1$, we get a deterministic reward $d$ if $a = L$ and $-d$ if $a = R$, where $d > 0$ is a positive constant. For states $s \in \{2,3\}$, irrespective of the chosen action, the reward is deterministically 1. Similarly, for states $s \in \{4,5\}$, the reward is deterministically 0. Now consider a Huber attack model where the attacker only perturbs rewards in state $s=1$ as follows: for state-action pair $(1, L)$ (resp., $(1, R))$, the observed reward is $d$ (resp., $-d$) with probability $1-\varepsilon$ and $-C$ (resp., $C$) with probability $\varepsilon$. Here, $C > 0$ is an attack signal that we will design carefully later. This immediately yields the following perturbed reward functions: $\tilde{R}_c(1,L) = (1-\varepsilon)d-\varepsilon C$ and $\tilde{R}_c(1,R)= -(1-\varepsilon)d+\varepsilon C.$ We now proceed to compute $Q^*$ and $\tilde{Q}^*_c$. First, observe that given our choice of attack and MDP model, $Q^*$ and $\tilde{Q}^*_c$ only differ at state $1$. Now using the Bellman optimality operators in~\eqref{eqn:Bellman} and~\eqref{eqn:perturb_Bellman}, it is easy to verify that $Q^*(s,a)=0$ for $s\in \{4,5\}, a \in \{L,R\}$, and $Q^*(s,a)=(1-\gamma p)^{-1}$ for $s\in \{2,3\}, a \in \{L,R\}$. Next, one can similarly check that 
\begin{equation}
\begin{aligned}
Q^*(1,L) &= d + \beta, \quad \tilde{Q}^*_c(1,L) = (1-\varepsilon) d - \varepsilon C + \beta \\
Q^*(1,R) &= -d + \beta, \quad \tilde{Q}^*_c(1,R) = - (1-\varepsilon) d + \varepsilon C + \beta,
\nonumber
\end{aligned}
\end{equation}
where $\beta = p \gamma (1-\gamma p)^{-1}.$ Now pick the corruption signal $C$ as $\left((2-\varepsilon)d + \kappa\right) \varepsilon^{-1}$, where $\kappa >0$ is a tunable parameter. With this choice, we  have $\tilde{Q}^*_c(1,L)= -d - \kappa + \beta$ and $\tilde{Q}^*_c(1,R)= d + \kappa + \beta$, yielding $\lVert \tilde{Q}_c^* - Q^* \rVert_\infty = \max_{a \in \mathcal{A}}\lvert \tilde{Q}_c^*(1,a) - Q^*(1,a) \rvert = 2d + \kappa.$ The claim of the theorem then follows from noting that $\kappa$ can be chosen arbitrarily by the attacker. Additionally, since $d, \kappa > 0$, it is not hard to see that under the attack constructed above, the optimal action at state $1$ also gets flipped: without the attack, the optimal action is $L$; under the attack, it is $R$.  
\end{proof}

Collectively, Theorems~\ref{thm:vulnerability} and~\ref{thm:example} reveal the vulnerability of the vanilla Q-learning algorithm to the Huber attack model. These findings directly motivate the robust algorithm we will develop in the next section. 
%-------------------- Algorithm -------------------- %
\section{\texorpdfstring{$\varepsilon$}--\texttt{Robust} Q-Learning algorithm}
\label{sec:Algorithm}
\begin{algorithm}[t]
\caption{Univariate Trimmed-Mean Estimator \cite{lugosi}}\label{alg:trim}
\begin{algorithmic}[1]
\Require Corrupted data set $Z_1, \ldots, Z_{M/2}$, $\tilde{Z}_1, \ldots \tilde{Z}_{M/2}$, corruption fraction $\varepsilon$, and confidence level $\delta$. 
\State Set $\zeta = 8 \varepsilon + 24\frac{\log(4/\delta)}{M}$.
\State Let $Z^*_1 \leq Z^*_2 \leq \cdots \leq  Z^*_{M/2}$ represent a non-decreasing arrangement of $\{Z_i\}_{i\in [M/2]}$. Compute quantiles: $\gamma = Z^*_{\zeta M/2}$ and $\beta=Z^*_{(1-\zeta) M/2}$.  
\State Compute robust mean estimate $\hat{\mu}_M$ as follows:
\begin{equation*}
   \hat{\mu}_M = \frac{2}{M}\sum_{i = 1}^{M/2} \phi_{\gamma, \beta}(\tilde{Z}_i);  
\phi_{\gamma, \beta}(x) = \begin{cases}
\beta & x>\beta\\
x & x \in [\gamma, \beta] \\ 
\gamma & x < \gamma
\end{cases} 
\end{equation*}
\end{algorithmic}
\end{algorithm}

Motivated by the vulnerability of vanilla Q-learning to reward-corruption attacks - as revealed in the previous section - we will develop a novel robust variant of the model-free synchronous Q-learning algorithm in this section. The basic idea behind our approach will be to use historical reward data collected for each state-action pair to construct a robust empirical Bellman operator in each iteration. To achieve this objective, we will drawn upon tools from the robust statistics literature; in particular, our algorithm will employ the univariate trimmed mean estimator developed in~\cite{lugosi} for robust mean estimation in the face of a strong-contamination model. Let us first briefly describe this estimator, and then explain how it can be used for our purpose. 

$\bullet$ \textbf{Univariate Trimmed-Mean Estimator.} The robust mean estimator in~\cite{lugosi} - outlined as Algorithm~\ref{alg:trim} - takes as input a confidence parameter $\delta$, a known corruption fraction $\varepsilon$, and a corrupted data set generated as follows. Consider a clean data set comprising of $M$ independent copies of a scalar random variable $Z$ with mean $\mu_Z$ and variance $\sigma^2_Z$. An adversary corrupts at most $\varepsilon M$ of these copies, and the resulting corrupted data set is then made available to the estimator. The estimation process involves splitting the corrupted data set into two equal parts, denoted by $Z_1, \ldots, Z_{M/2}$, $\tilde{Z}_1, \ldots \tilde{Z}_{M/2}$. One of the parts is used to compute certain quantile levels for filtering out extreme values (line 2 of Algorithm~\ref{alg:trim}). The robust estimate $\hat{\mu}_Z$ of $\mu_Z$ is obtained by averaging the data points in the other part, with those data points falling outside the estimated quantile levels truncated prior to averaging (line 3 of Algorithm~\ref{alg:trim}). {In what follows, we will use $\texttt{TRIM}[ \{Z_i\}_{i \in [M]}, \varepsilon, \delta\}]$ to succinctly describe the output of Algorithm~\ref{alg:trim}.} 

The following result characterizes the performance of Algorithm \ref{alg:trim}, and will be invoked by us in our subsequent analysis.
\begin{tcolorbox}
\begin{theorem} \cite[Theorem 1]{lugosi} 
\label{thm:lugosi}
Consider the trimmed mean estimator in Algorithm \ref{alg:trim}. Suppose $\varepsilon \in [0,1/16)$, and let $\delta \in (0,1)$ be such that $\delta \geq 4 e^{-M/2}$. Then, there exists an universal constant $C$, such that with probability at least $1-\delta$,
\begin{equation}
|\hat{\mu}_Z - \mu_Z| \leq C\sigma_Z\left(\sqrt{\varepsilon}+\sqrt{\frac{\log(4/\delta)}{M}} \right).
\label{eqn:lugosi_bnd}
\end{equation}
\end{theorem}
\end{tcolorbox}

We now describe how the estimator in Algorithm~\ref{alg:trim} can be used to generate robust estimates of the optimal Q function. 

$\bullet$ $\varepsilon$-\textbf{Robust Q-Learning Algorithm.} Our proposed robust Q-learning algorithm - outlined in Algorithm~\ref{algo:algo2} - starts with an initial Q-function estimate $Q_0$, and takes as input a corruption fraction $\varepsilon$, and a failure probability $\delta \in (0,1).$ In each iteration $t$, for each $(s,a) \in \mc{S} \times \mc{A}$, the learner gets to observe $s_t(s,a)$ drawn independently from $P(\cdot| s,a)$ (as in the standard synchronous Q-learning setting), and a potentially corrupted version of $r_t(s,a) \sim \mc{R}(s,a)$ denoted by $y_t(s,a)$. Here, the corruption adheres to the strong contamination reward attack model specified in Section~\ref{sec:prob_form}. Our strategy to safeguard against adversarial reward contamination is twofold: \emph{reward-filtering} and \emph{thresholding}. We describe these ideas below. 

\textbf{Reward-Filtering}: First, instead of directly using $y_t(s,a)$ to update $Q_t(s,a)$, we instead compute a robust estimate of the expected reward $R(s,a)$ for state-action pair $(s,a)$ by using the univariate trimmed mean estimator in Algorithm~\ref{alg:trim}. This is done by invoking Algorithm~\ref{alg:trim} with (i) data set $\{y_k(s,a)\}_{0 \leq k \leq t}$ comprising of the history of reward observations for $(s,a)$, (ii) the corruption fraction $\varepsilon$, and (iii) a finer (relative to $\delta$) confidence parameter $\delta_1=\delta/(2 |\mc{S}| |\mc{A}| T)$. We denote the operation carried out by Algorithm~\ref{alg:trim} succinctly by using the \texttt{TRIM} function in line 6 of Algorithm~\ref{algo:algo2}. Let the output of this operation be denoted $\tilde{r}_t(s,a)$. While it is tempting to use $\tilde{r}_t(s,a)$ to update $Q_t(s,a)$, one needs to exercise caution as we explain next. 

\textbf{Reward-Thresholding}. From Theorem~\ref{thm:lugosi}, a couple of points are worth noting about the  guarantee afforded by Algorithm~\ref{alg:trim}. First, the guarantee holds only if the number of data samples exceeds $T_{lim}$, where 
 \begin{equation}
    T_{lim} \triangleq \ceil{2 \log\left({4}/{\delta_1}\right)}. 
    \label{eqn:Tlim}
 \end{equation}

 \begin{algorithm}[t]
\caption{The \texttt{$\varepsilon$-Robust} Q-Learning Algorithm}
\label{algo:algo2}
   \begin{algorithmic}[1]
\State \textbf{Input:} Initial estimate $Q_0 \in \mathbb{R}^{|\mathcal{S}||\mathcal{A}|}$, step-size $\alpha$, corruption fraction $\varepsilon$, confidence level $\delta$, total iterations $T$
\For {$t=0,1, \ldots T$}
\For {$(s,a) \in \mc{S} \times \mc{A}$}
\State Sample $s_t(s,a) \sim P(\cdot|s,a)$.
\State Observe $y_t(s,a)$.
\State $\tilde{r}_t(s,a) \leftarrow \texttt{TRIM}\left[\{y_k(s,a)\}_{0 \leq k \leq t}, \varepsilon, \delta_1\right],$
where $\delta_1 = \delta/(2 |\mc{S}| |\mc{A}| T).$
\If{$\lvert \tilde{r}_t(s,a)\rvert > G_t$}
\State Set $\tilde{r}_t(s,a) \leftarrow \texttt{SIGN}(\tilde{r}_t(s,a)) \times G_t$, where $G_t$ is the threshold function in Eq.~\eqref{eqn:Gt}. 
\EndIf
\State Update $Q_t(s,a)$ as per Eq.~\eqref{eqn:eqnrobust1}.
\EndFor 
\EndFor
\end{algorithmic} 
\end{algorithm}

Second, the guarantee is not deterministic; instead, it only holds with high probability. For technical reasons that will become apparent later in our analysis, we need the sequence $\{Q_t\}$ of iterates generated by our algorithm to be uniformly bounded deterministically. However, the above discussion suggests that simply using the output of the \texttt{TRIM} function to perform updates will not suffice to achieve this goal. As such, we employ a second layer of safety by carefully defining a threshold function as follows: 
\begin{equation}\label{eqn:Gt}
\begin{aligned}
G_t=\begin{cases}
      2 \R, & \text{$0 \leq t \leq T_{lim}$}\\
      C \R \left(\sqrt{\frac{\log\left({\frac{4}{\delta_1}}\right)}{t}}+ \sqrt{\varepsilon}\right)+\R, & \text{$t \ge T_{lim}+1$}\\
    \end{cases} 
\end{aligned}
\end{equation}
where $C$ is the universal constant in Eq.~\eqref{eqn:lugosi_bnd}. 

Whenever the output of the \texttt{TRIM} function exceeds $G_t$ in magnitude, we control it via the thresholding operation in line 8 of Algorithm~\ref{algo:algo2}. This ensures boundedness of iterates. When adequate data samples have not been collected, i.e., $ t\leq T_{lim}$, we cannot rely on the statistical bound from Theorem~\ref{thm:lugosi}; hence, we use the crude bound $\R$ on the rewards to design $G_t$. However, once enough samples have in fact been collected, we would expect the output of \texttt{TRIM} to concentrate around $R(s,a)$ with high-probability. This motivates the choice of $G_t$ for $t \geq T_{lim}+1$ based on Eq.~\eqref{eqn:lugosi_bnd}. As we argue later in Lemma~\ref{lemma:adversarialbnd}, the above choice of $G_t$ ensures that for $t \geq T_{lim}+1$, the condition in line 7 of Algorithm~\ref{algo:algo2} gets violated with high probability, and $\tilde{r}_t(s,a)$ remains the output of the \texttt{TRIM} function instead of the more conservative estimate in line 8. This turns out to be crucial for achieving tight rates. After filtering and thresholding, $Q_t(s,a)$ is updated as 
\begin{equation}
\label{eqn:eqnrobust1}
\begin{aligned}
Q_{t+1}(s,a) &= (1- \alpha)Q_t(s,a) + \alpha \left(\tilde{r}_t(s,a) + \gamma \max_{a'\in \mathcal{A}}Q_t(s_t(s,a),a')\right). 
\end{aligned}
\end{equation}

This completes the description of Algorithm~\ref{algo:algo2}. To state our main result concerning its finite-time performance, let us define $d_t \triangleq \lVert Q_{t} - Q^{*}\rVert_\infty$. Our main result is as follows.
\begin{tcolorbox}
\begin{theorem}\label{theorem:theoremmainresult} (\textbf{Robust Q-learning bound}) Suppose the corruption fraction satisfies $\varepsilon \in [0,1/16)$. Then, given any $\delta \in (0,1)$, the output of Algorithm~\ref{algo:algo2} with step-size $\alpha = \frac{\log T}{(1-\gamma)T}$  satisfies the following bound with probability at least $1-\delta$:
\begin{equation}
d_T \leq \frac{d_0}{T} + O\left( \frac{\R}{(1-\gamma)^{\frac{5}{2}}}  \frac{\log T}{\sqrt{T}} \sqrt{ \log \left(\frac{|\mathcal{S}||\mathcal{A}|T}{\delta}\right)} +  \frac{\R\sqrt{\varepsilon}}{1-\gamma}\right). 
\label{eqn:main_conv_bnd}
\nonumber
\end{equation}
\end{theorem}
\end{tcolorbox}
We defer the proof of this result to Section~\ref{sec:proof}. 

\textbf{Discussion.} To parse the guarantee in Theorem~\ref{theorem:theoremmainresult}, let us simplify the bound in the theorem to obtain 
\begin{equation}
d_T \leq \frac{d_0}{T}+\underbrace{\tilde{O}\left( \frac{\bar{r}}{(1-\gamma)^{\frac{5}{2}} \sqrt{T}}\right)}_{T_1} + \underbrace{O\left( \frac{\R \sqrt{\varepsilon}}{1-\gamma} \right)}_{T_2}. 
\end{equation}
From the above display, we note that our proposed Robust Q-Learning algorithm yields a high-probability $\ell_{\infty}$ sample-complexity bound that features two dominant terms: term $T_1$ captures the behavior of the algorithm in the absence of attacks, while term $T_2$ captures the effect of reward corruption. The dependence of term $T_1$ on both $1/(1-\gamma)$ and the number of iterations $T$ exactly matches that in the recent works~\cite{Waiwright} and~\cite{Qu}. Thus, in the absence of attacks, our result is consistent with prior art. In the presence of strong-contamination attacks, our algorithm guarantees convergence to $Q^*$ up to an additive error that scales as $O(\bar{r} \sqrt{\varepsilon})$. We conjecture that such a term is unavoidable. To see why, consider a trivial MDP that comprises of just one state $s$ and one action $a$. In this case, the problem of learning $Q^*$ essentially boils down to estimating the mean $R(s,a)$ using a set of noisy and corrupted data points. With this reduction, we can now invoke fundamental lower bounds from the robust mean-estimation literature which tell us that an error of $\Omega(\sigma \sqrt{\varepsilon})$ is unavoidable~\cite{lugosi, Diakonikolas_Kane_2023} when the clean noise samples have variance $\sigma^2$ . Since the noisy rewards in our setting come from the interval $[-\R, \R]$, $\R^2$ is precisely the variance-proxy for our setting. While the above argument can be formalized, we omit it here due to space constraints. \emph{To sum up, our work provides the first near-optimal guarantee for Q-learning under a strong reward corruption model.}  

{
\begin{remark}
Notice that our proposed algorithm uses knowledge of the bound $\bar{r}$ on the rewards to design the dynamic threshold in Eq.~\eqref{eqn:Gt}. At this stage, one might consider a much simpler algorithm that simply ignores rewards that have magnitude larger than $\bar{r}$. Unfortunately, such a strategy is not guaranteed to work when the true reward distributions have infinite support. However, in Section~\ref{sec:unboundedrewards}, we will explain that with minor modifications to Algorithm~\ref{algo:algo2}, one can continue to handle reward distributions with infinite support, under the assumption of bounded second moments. 
\end{remark}
} 
%------------ Main Analysis ----------------%
\section {Proof of Theorem~\ref{theorem:theoremmainresult}}\label{sec:proof}
In this section, we will provide a detailed proof of our main result, namely Theorem~\ref{theorem:theoremmainresult}. There are two sources of error in our update rule: one due to the randomness (i.e., noise) that originates from the sampling process, and the other due to reward-poisoning by the adversary. As such, the first step in our analysis is to set up an error-decomposition that clearly separates out the statistical  terms in our update rule from those originating due to adversarial contamination. To that end, with some simple algebra, observe that our proposed Robust Q-learning update rule in Eq.~\eqref{eqn:eqnrobust1} takes the following form:
     \begin{equation}\label{eqn:iteratedeqn}
     \begin{aligned}
          & Q_{t+1}(s,a) = (1- \alpha)Q_t(s,a) \\
          & + \alpha \underbrace{\left(R(s,a) +\gamma \mathbb{E}_{s' \sim P(\cdot | s,a)}\left[\max_{a' \in \mathcal{A}} Q_t(s',a')\right]\right)}_{(\mathcal{T}^* Q_t)(s,a)} + \alpha\underbrace{ [\tilde{r}_t(s,a)- R(s,a)]}_{\mathcal{E}_{t}(s,a)}\\
          & + \alpha\underbrace{ \left(\gamma \max_{a'\in \mathcal{A}}Q_t(s_t(s,a),a')- \gamma \mathbb{E}_{s' \sim P(\cdot | s,a)}\left[\max_{a' \in \mathcal{A}} Q_t(s',a')\right]\right)}_{\mathcal{D}_{t}(s,a)},\\
     \end{aligned}
    \end{equation}
where we used the definition of the Bellman operator in Eq.~\eqref{eqn:Bellman}. Stacking up the components $Q_t(s,a)$, $\mathcal{E}_{t}(s,a)$, and $\mathcal{D}_{t}(s,a)$ for each state-action pair $(s,a) \in \mathcal{S} \times \mathcal{A}$ into vectors $Q_t, \mathcal{E}_t$, and $\mathcal{D}_t \in \mathbb{R}^{|\mathcal{S}| |\mathcal{A}|}$, respectively, the display in Eq.~\eqref{eqn:iteratedeqn} takes the following compact form: 
\begin{equation}\label{eqn:Iteration_Vector}
    Q_{t+1} = (1-\alpha)Q_t + \alpha \left(\mathcal{T}^* (Q_t)  +\mathcal{E}_{t}+\mathcal{D}_{t}\right).
\end{equation}
Now subtracting $Q^*$ from both sides of Eq.~\eqref{eqn:Iteration_Vector}, using $\mathcal{T}^* (Q^*) =Q^*$, and rolling out the resulting equation yields: 
\begin{equation}
    \begin{aligned}
    Q_{t+1}-Q^* & = (1-\alpha)^{t+1}(Q_0-Q^*) \\
        & \hspace{-6mm} + \sum_{k=0}^{t} \alpha (1-\alpha)^{t-k} (\mathcal{T}^* (Q_k) - \mathcal{T}^* (Q^*))+\sum_{k=0}^{t} \alpha (1-\alpha)^{t-k} \mathcal{E}_{k} \\
        & \hspace{-6mm} + \sum_{k=0}^{t} \alpha (1-\alpha)^{t-k} \mathcal{D}_{k}.
    \end{aligned}
\label{eqn:unrolled}
\end{equation}
Defining $d_t \triangleq \lVert Q_{t} - Q^{*}\rVert_\infty, \forall t \geq 0,$ and taking the infinity norm on both sides of Eq.~\eqref{eqn:unrolled}, we obtain:
\begin{equation} \label{eqn:Final_eqn}
        d_{t+1} \le (1-\alpha)^{t+1}d_0 + \gamma \sum_{k=0}^{t} \alpha (1-\alpha)^{t-k} d_k + A_{1,t} + A_{2,t}, 
\end{equation}    
where 
\begin{equation}
        A_{1,t} \triangleq \lVert \sum_{k=0}^{t} \alpha (1-\alpha)^{t-k} \mathcal{D}_{k}\rVert_{\infty}, A_{2,t} \triangleq \lVert \sum_{k=0}^{t} \alpha (1-\alpha)^{t-k} \mathcal{E}_{k} \rVert_{\infty}.
\label{eqn:error_terms}
\end{equation}
To arrive at Eq.~\eqref{eqn:Final_eqn}, we used the contraction property of the Bellman operator in Eq.~\eqref{eqn:Bellmancontraction} to infer that 
  $\lVert \mathcal{T}^* (Q_k) - \mathcal{T}^* (Q^*)\rVert_{\infty}\le \gamma\lVert Q_k - Q^*\rVert_{\infty} \leq \gamma d_k.$ We have thus proved the following lemma. 

\begin{lemma} (\textbf{Error-Decomposition}) 
\label{lemma:errordec}
The robust update rule in Eq.~\eqref{eqn:eqnrobust1} satisfies the error-bound in Eq.~\eqref{eqn:Final_eqn} for all $t\geq 0.$ 
\end{lemma}

The remainder of our analysis will focus on providing bounds for the terms $A_{1,t}$ and $A_{2,t}$ featuring in Eq.~\eqref{eqn:error_terms}. 

\noindent $\bullet$ \textbf{Bounding the Effect of Noise.} We first turn our attention to bounding the term $A_{1,t}$ in Eq.~\eqref{eqn:Final_eqn}. To that end, let $\mathcal{F}_t$ denote the $\sigma$-algebra generated by $\{Q_k\}_{0 \leq k \leq t}$. Now for each $(s,a) \in \mathcal{S} \times \mathcal{A}$, since $s_t(s,a)$ is generated independently from $P(\cdot|s,a),$ and $Q_t$ is $\mathcal{F}_t$-adapted, it is easy to see that $\mathbb{E}[\mathcal{D}_t(s,a)|\mathcal{F}_t]=0.$ Thus, $\sum_{k=0}^{t} \alpha (1-\alpha)^{t-k} \mathcal{D}_{k}$ has a martingale structure that we can hope to exploit by appealing to the Azuma-Hoeffding bound for martingales. However, this requires care, as we describe next. To apply Azuma-Hoeffding in its standard form, we need to argue that the martingale difference sequence $\{\mc{D}_t(s,a)\}$ is uniformly bounded with probability $1$. This difference sequence features $Q_t$, which, in turn, contains the filtered reward $\tilde{r}_t(s,a)$; see Eq.~\eqref{eqn:eqnrobust1}. Since $\tilde{r}_t(s,a)$ depends on the sequence of observations $\{y_k(s,a)\}_{0 \leq k \leq t}$, and some of these observations can be arbitrarily corrupted by the adversary, we need to carefully justify the boundedness of $\mathcal{D}_t(s,a).$ \emph{Unfortunately, Theorem~\ref{thm:lugosi} cannot help us in this regard since it only provides a bound that holds with high-probability; however, we seek a bound that holds almost surely}. As revealed by our next result, this is precisely where the additional robustness afforded by the trimming step in line 8 of Algorithm~\ref{algo:algo2} will play a crucial role. To simplify the analysis, we will assume without loss of generality that $Q_0=0.$

\begin{lemma} (\textbf{Boundedness of Iterates}) \label{lemma:uniformbnd} The following is true for the iterates generated by Algorithm~\ref{algo:algo2}: 
$$ \lvert Q_{t}(s,a) \rvert \le \frac{3C\R}{1-\gamma}, \forall (s,a) \in \mathcal{S} \times \mathcal{A}, \forall t \geq 0, $$
where $C$ is the universal constant in Eq.~\eqref{eqn:Gt}.
    \begin{proof}
We will prove this result via induction. Fix any $(s,a) \in \mc{S} \times \mc{A}$. Since $Q_0=0$, the claim of the lemma holds trivially at $t=0$. This completes the base case of induction. Now suppose the bound claimed in the lemma holds for all $0 \leq k \leq t.$ To show that it also holds at iteration $t+1$, we first need an estimate on $|\tilde{r}_t(s,a)|$. From lines 7 and 8 of Algorithm~\ref{algo:algo2}, it is evident that $|\tilde{r}_t(s,a)| \leq G_t$, where $G_t$ is the trimming radius in Eq.~\eqref{eqn:Gt}. Thus, to control $|\tilde{r}_t(s,a)|$, it suffices to bound $G_t$. Now from the definition of $T_{lim}$ in Eq.~\eqref{eqn:Tlim}, and the fact that $\varepsilon \in (0,1)$, it is easy to see from Eq.~\eqref{eqn:Gt} that $G_t \leq 3C \R, \forall t \geq T_{lim}+1$; here, we used $C \geq 1$. Since $G_t = 2\R$ for $ 0 \leq t \leq T_{lim}$, we conclude $|\tilde{r}_t(s,a)| \leq 3C \R, \forall t \geq 0$. To proceed with the induction step, let us now observe from Eq.~\eqref{eqn:eqnrobust1} that 
        \begin{equation}
            \begin{aligned}
               & \lvert Q_{t+1}(s,a) \rvert \le (1 - \alpha) \lvert Q_{t}(s,a) \rvert + \alpha \rvert \tilde{r}_t(s,a) + \gamma \max_{a'\in \mathcal{A}}Q_t(s_t(s,a),a')\lvert\\
               & \overset{(\texttt{a})}{\le}(1 - \alpha) \lvert Q_{t}(s,a) \rvert + \alpha \left(\rvert \tilde{r}_t(s,a)\lvert + \gamma \max_{a'\in \mathcal{A}} \rvert Q_t(s_t(s,a),a')\lvert \right) \\
               & \overset{(\texttt{b})}{\le} (1 -\alpha) \frac{3C\R}{1-\gamma} + \alpha \left( 3C\R + \gamma \frac{3C\R}{1-\gamma}\right) \\
               & = (1 -\alpha) \frac{3C\R}{1-\gamma} + \alpha \frac{3C\R}{1-\gamma} = \frac{3C\R}{1-\gamma},\\
            \end{aligned}
        \end{equation}
where for (a), we used the triangle inequality, and for (b), we used the induction hypothesis in tandem with the previously established fact that $|\tilde{r}_t(s,a)| \leq 3C \R, \forall t \geq 0$. This completes the induction step and the proof. 
       \end{proof}
\end{lemma}

Armed with Lemma~\ref{lemma:uniformbnd}, we can now appeal to the Azuma-Hoeffding Theorem for martingales to control the noise-induced term $A_{1,t}.$ To do so, we note that a sequence of random variables $Z_1, Z_2, Z_3, \ldots$ is called a \emph{martingale difference sequence} w.r.t. some filtration $\mc{G}_t$ if, for each $t$, it holds that $\mathbb{E}[Z_t| \mc{G}_{t-1}]=0.$ To keep the paper self-contained, we recall the following version of Azuma-Hoeffding from~\cite{chungconc}. 

\begin{lemma} (\textbf{Azuma-Hoeffding})
\label{lemma:AHinequality}
Let $Z_1, Z_2, Z_3, \ldots$ be a martingale difference sequence with $|Z_i| \le c_i$ for all $i \in \mathbb{N},$ where each $c_i$ is a positive real. Then, for all $\lambda \ge 0$:
\[
\mathbb{P}\left(\left|\sum_{i=1}^{n} Z_i\right| \ge \lambda\right) \le 2e^{-\frac{\lambda^2}{2\sum_{i=1}^{n}{c}_i^2}}.
\]
\end{lemma}

We have the following result that controls the effect of noise in our update rule. 

\begin{lemma}
\label{lemma:noise} (\textbf{Noise Bound}) With probability at least $1- \delta/2$, the following bound holds simultaneously $\forall t \in [T]$: 
    \begin{equation}
\lVert \sum_{k=0}^{t} \alpha (1-\alpha)^{t-k} \mathcal{D}_{k}\rVert_\infty \le \frac{6C\R\gamma}{1-\gamma} \sqrt{2{\alpha} \log\left(\frac{4|\mathcal{S}||\mathcal{A}|T}{\delta}\right)},
\nonumber
\end{equation}
where $\mathcal{D}_{k}$ is as defined in Eq.~\eqref{eqn:iteratedeqn}.
    \begin{proof}
Let us fix a state-action pair $(s,a) \in \mc{S} \times \mc{A}$, and a time-step $t \in [T]$. We have already argued that $\mathbb{E}[\mathcal{D}_t(s,a)|\mathcal{F}_t]=0$, i.e., $\{\mc{D}_k(s,a)\}$ is a martingale difference sequence. Furthermore, from Lemma~\ref{lemma:uniformbnd}, we have 
\begin{equation}
\begin{aligned}
\label{eqn:MDS}
    \lvert \mathcal{D}_{t}(s,a)\rvert &= \lvert \gamma \max_{a' \in \mathcal{A}} Q_t(s_t(s,a), a') - \gamma \mathbb{E}_{s' \sim P(\cdot | s, a)} \max_{a' \in \mathcal{A}} Q_t(s', a') \rvert \\
    & \le \gamma \max_{a' \in \mathcal{A}} \lvert  Q_t(s_t(s,a), a') \rvert + \gamma \mathbb{E}_{s' \sim P(\cdot | s, a)} \max_{a' \in \mathcal{A}} \vert Q_t(s', a') \rvert\\
    & \leq \frac{6C\R \gamma}{1-\gamma} \triangleq \Delta.
   \end{aligned} 
   \nonumber
\end{equation}
Noting that $\{ \alpha (1-\alpha)^{t-k}\mc{D}_k(s,a)\}$ is also a martingale difference sequence, and applying Lemma~\ref{lemma:AHinequality}, we note that the following bound holds with probability at least $1-\delta_2$: 
        \begin{equation}
        \begin{aligned}
        \left|\sum_{k=0}^{t}\alpha(1-\alpha)^{t-k} \mathcal{D}_{k}(s,a)\right| &\leq \Delta \sqrt{ 2 \alpha^2 \log\left(\frac{2}{\delta_2}\right) \sum_{k=0}^t (1-\alpha)^{2(t-k)}}\\
        &\le \Delta \sqrt{ 2 \alpha^2 \log\left(\frac{2}{\delta_2}\right) \sum_{k=0}^t (1-\alpha)^{(t-k)}}\\
        &\le \Delta \sqrt{ 2 \alpha^2 \log\left(\frac{2}{\delta_2}\right) \sum_{i=0}^{\infty} (1-\alpha)^{i}}\\
        & = \Delta \sqrt{ 2 \alpha \log\left(\frac{2}{\delta_2}\right)},
        \end{aligned}
\nonumber
        \end{equation}
where for the second inequality, we used $(1-\alpha) < 1.$ Using the above fact, and union-bounding over all $(s,a) \in \mc{S} \times \mc{A}$, we conclude that the following bound holds with probability at least $1-\delta_2|S| |A|$: 
        \begin{equation}\label{eqn:MDbnd}
        \begin{aligned}
               \lVert \sum_{k=0}^{t} \alpha (1-\alpha)^{t-k} \mathcal{D}_{k}\rVert_\infty &= \max_{{(s,a) \in \mathcal{S} \times \mathcal{A}}}\lvert \sum_{k=0}^{t} \alpha (1-\alpha)^{t-k} \mathcal{D}_{k}(s,a)\rvert \\
               &\le \Delta \sqrt{ 2 \alpha \log\left(\frac{2}{\delta_2}\right)}. 
        \end{aligned}
        \end{equation}
Union-bounding over all $t\in [T]$, we conclude that the bound in Eq.~\eqref{eqn:MDbnd} holds simultaneously for all $t \in [T]$ with probability at least $1-\delta_2 |\mathcal{S}||\mathcal{A}| T$. The claim of the lemma now follows by setting $\delta_2 = \delta/(2|\mathcal{S}||\mathcal{A}| T).$        
\end{proof}
\end{lemma}

With the above developments, we now have a handle over the term $A_{1,t}$ in the main error bound of Eq.~\eqref{eqn:Final_eqn}. 
\vspace{2mm}
\\
\noindent $\bullet$ \textbf{Bounding the Effect of Adversarial Contamination.} The effect of adversarial contamination gets manifested in the term $A_{2,t}$ of Eq.~\eqref{eqn:Final_eqn}. The following key result helps us control this term. 

\begin{lemma} \label{lemma:adversarialbnd} (\textbf{Adversarial Corruption Bound}) Suppose $\varepsilon \in [0,1/16)$. Then, with probability at least $1- \delta/2$, the following bound holds simultaneously $\forall t \in [T]$:
\begin{equation}
\lVert \sum_{k=0}^{t} \alpha (1-\alpha)^{t-k} \mathcal{E}_{k}\rVert_\infty \le 8 \alpha C \R \sqrt{T  \log\left(\frac{8|\mathcal{S}||\mathcal{A}|T}{\delta}\right)}+C\R\sqrt{\varepsilon},
\nonumber
\end{equation}
where $\mathcal{E}_{k} $ is as defined in Eq.~\eqref{eqn:iteratedeqn}.
\begin{proof} We will split our analysis into two separate cases.\\
\textbf{Case I}. Consider first the case when $t \in [T_{lim}]$. Fix any $(s,a) \in \mc{S} \times \mc{A}.$ Recalling that $\mathcal{E}_{t}(s,a) = \tilde{r}_t(s,a)- R(s,a)$, we have
$$ |\mc{E}_t(s,a)| \leq |\tilde{r}_t(s,a)| + |R(s,a)| \leq G_t + \R \leq 3\R,$$
where we used $|\tilde{r}_t(s,a)| \leq G_t, \forall t \geq 0$ (based on lines 7 and 8 of Algorithm~\ref{algo:algo2}), and Eq.~\eqref{eqn:Gt}. We also used the fact that since the uncorrupted reward $r_t(s,a) \in [-\R, \R]$, the expected value of $r_t(s,a)$, namely $R(s,a)$, can have magnitude no larger than $\R.$ Thus, for $t \in [T_{lim}]$, we have 
\begin{equation}
\begin{aligned}
\left|\sum_{k=0}^{t}\alpha(1-\alpha)^{t-k} \mathcal{E}_{k}(s,a)\right| & \leq \sum_{k=0}^{t}\alpha(1-\alpha)^{t-k} \left|\mathcal{E}_{k}(s,a)\right|\\
&\leq 3 \alpha \R  \sum_{k=0}^{t}(1-\alpha)^{t-k} \\
&\leq 3 \alpha \R \sum_{k=0}^{T_{lim}}(1-\alpha)^{t-k} \\
&\leq 3 \alpha \R T_{lim}. 
\end{aligned}
\end{equation}
Since the above argument applies identically to every $(s,a) \in \mc{S} \times \mc{A}$, we conclude that for every $t \in [T_{lim}]$, 
\begin{equation}\label{eqn:Errbnd}
            \begin{aligned}
               \lVert \sum_{k=0}^{t} \alpha (1-\alpha)^{t-k} \mathcal{E}_{k}\rVert_\infty &= \max_{{(s,a) \in \mathcal{S} \times \mathcal{A}}}\lvert \sum_{k=0}^{t} \alpha (1-\alpha)^{t-k} \mathcal{E}_{k}(s,a)\rvert \\
               &\le 3 \alpha \R T_{\lim}. 
            \end{aligned}
        \end{equation}
Note that the above bound holds deterministically. 

\textbf{Case II}. Now suppose $ T_{lim}+1 \leq t \leq T$, and fix a $(s,a) \in \mc{S} \times \mc{A}$ as before. To control $\mc{E}_t(s,a)$ in this case, we wish to use the bound from robust mean estimation in Theorem~\ref{thm:lugosi}. To that end, we make the following observations. First, under our synchronous sampling model, the reward samples $\{r_k(s,a)\}_{0 \leq k \leq t}$ are i.i.d. random variables with mean $R(s,a)$. Second, since $r_t(s,a) \in [-\R, \R], \forall t\geq 0$, the variance of these samples is at most $\R^2.$ Third, under the assumptions on our attack model, at most $\varepsilon$-fraction of the samples in $\{r_k(s,a)\}_{0 \leq k \leq t}$ can be corrupted by the adversary. Thus, invoking Theorem~\ref{thm:lugosi}, we note that the output $\tilde{r}_t(s,a)$ of the \texttt{TRIM} function in line 6 of Algorithm~\ref{algo:algo2} satisfies the following bound with probability at least $1-\delta_1$:
\begin{equation}
|\tilde{r}_t(s,a) - R(s,a)| \leq  \underbrace{C \R \left(\sqrt{\frac{\log\left({\frac{4}{\delta_1}}\right)}{t}}+ \sqrt{\varepsilon}\right)}_{(*)}. 
\label{eqn:robustmeanbnd}
\end{equation}
Now consider an event $\mc{X}$ where the above bound holds simultaneously for all $(s,a) \in \mc{S} \times \mc{A}$, and for all $T_{lim}+1 \leq t \leq T$. Union-bounding over all state-action pairs and time-steps in the above interval, we note that event $\mc{X}$ has measure at least $1- \delta_1 |\mc{S}| |\mc{A}| (T-T_{lim}) > 1 - \delta_1 |\mc{S}| |\mc{A}| T.$ For the remainder of the analysis, we will condition on the event $\mc{X}$. Since $|R(s,a)| \leq \R$, observe that given our choice of $G_t$ in Eq.~\eqref{eqn:Gt}, the output $\tilde{r}_t(s,a)$ of the \texttt{TRIM} function satisfies $|\tilde{r}_t(s,a)| \leq G_t$ on event $\mc{X}$. Thus, the condition in line 7 of Algorithm~\ref{algo:algo2} gets violated and line 8 gets bypassed. In other words, $\tilde{r}_t(s,a)$ remains as in line 6 of Algorithm~\ref{algo:algo2}, and we obtain that $|\mc{E}_t(s,a)|$ is bounded from above by $(*)$ on event $\mc{X}$, implying 
    \begin{equation}
    \begin{aligned}
      &\lvert \sum_{k=T_{lim}+1}^{t} \alpha (1-\alpha)^{t-k} \mathcal{E}_{k}(s,a)\rvert  \le \sum_{k=T_{lim}+1}^{t} \alpha (1-\alpha)^{t-k} \lvert \mathcal{E}_{k}(s,a) \rvert \\
      & \overset{(a)}{\le} C \R \sum_{k=T_{lim}+1}^{t} \alpha (1-\alpha)^{t-k} \left( \sqrt{\frac{\log\left(\frac{4}{\delta_1}\right)}{k}} + \sqrt{\varepsilon} \right)\\
      & \le \alpha C \R \sqrt{\log\left(\frac{4}{\delta_1}\right)} \sum_{k=T_{lim}+1}^{t} \frac{1}{\sqrt{k}} +  \alpha C \R \sum_{k=0}^{\infty} (1-\alpha)^{k} \sqrt{\varepsilon} \\
      &    \overset{(b)}{\le} {2\alpha C \R \sqrt{T \log\left(\frac{4}{\delta_1}\right)}} +  C\R \sqrt{\varepsilon}.
    \end{aligned}
    \end{equation}
For (a), we used $|\mc{E}_t(s,a)| \leq (*)$ on event $\mc{X}$, and for (b), we used a standard trick of bounding a sum by an integral to control the first term in the inequality. Combined with our analysis for Case 1, we then have on the event $\mc{X}$:
\begin{equation}
\begin{aligned}
\left|\sum_{k=0}^{t}\alpha(1-\alpha)^{t-k} \mathcal{E}_{k}(s,a)\right| & \leq \left|\sum_{k=0}^{T_{lim}}\alpha(1-\alpha)^{t-k} \mathcal{E}_{k}(s,a)\right|\\
&+\left|\sum_{k={T_{lim}+1}}^{t}\alpha(1-\alpha)^{t-k} \mathcal{E}_{k}(s,a)\right|\\
& \hspace{-15mm} \leq 3\alpha \bar{r} T_{lim} + { 2\alpha C \R \sqrt{T \log\left(\frac{4}{\delta_1}\right)}} +  C\R \sqrt{\varepsilon}\\
& \hspace{-15mm} \overset{(a)}{\leq} 3\alpha \bar{r} \sqrt{T_{lim}}{\sqrt{T}} + {2\alpha C \R \sqrt{T \log\left(\frac{4}{\delta_1}\right)}} +  C\R \sqrt{\varepsilon}\\
& \hspace{-15mm} \overset{(b)}\leq {8 \alpha C \R \sqrt{T \log\left(\frac{4}{\delta_1}\right)}} +  C\R \sqrt{\varepsilon},
\end{aligned}
\label{eqn:Err_bnd_final}
\end{equation}
where for (a), we used $T_{lim} \leq T$, and for (b), we used the expression for $T_{lim}$ in Eq.~\eqref{eqn:Tlim}, and simplified using $C\geq 1.$ We conclude that on the event $\mc{X}$, the bound in Eq.~\eqref{eqn:Err_bnd_final} applies simultaneously to all $t \in [T]$, and all $(s,a) \in \mc{S} \times \mc{A}.$ The claim of the lemma now follows immediately by picking $\delta_1 = \delta/(2 |\mc{S}| |\mc{A}| T).$
\end{proof}
\end{lemma}

We are now ready to complete the proof of Theorem~\ref{theorem:theoremmainresult}.
\begin{proof} (\textbf{Proof of Theorem~\ref{theorem:theoremmainresult})} Our claim is that with probability at least $1-\delta$, the following bound holds simultaneously $\forall t\in [T]\cup \{0\}:$
\begin{equation}
d_t \le (1-\alpha(1-\gamma))^t d_0 + \frac{W}{1-\gamma}, \hspace{1mm} \textrm{where} 
\label{eqn:ind_claim}
\end{equation}
\begin{equation} 
\resizebox{0.7\hsize}{!}{$
W \triangleq \frac{6C\R\gamma}{1-\gamma} \sqrt{2{\alpha} \log\left(\frac{4|\mathcal{S}||\mathcal{A}|T}{\delta}\right)} + 8 \alpha C \R \sqrt{T  \log\left(\frac{8|\mathcal{S}||\mathcal{A}|T}{\delta}\right)}+C\R\sqrt{\varepsilon}$}.
\label{eqn:peturb_bnd}
\end{equation}
We remind the reader here that $d_t = \lVert  Q_t - Q^* \rVert_{\infty}$. We will prove the above claim via induction. Trivially, the induction claim holds deterministically for $t=0$. Now from Lemmas \ref{lemma:noise} and \ref{lemma:adversarialbnd}, let us note that there exists an event - say $\mc{Y}$ - of measure at least $1-\delta$ on which $A_{1,t}+A_{2,t} \leq W, \forall t \in [T]$, where $A_{1,t}$ and $A_{2,t}$ are as defined in Eq.~\eqref{eqn:error_terms}. We will condition on the ``good" event $\mc{Y}$ for the remainder of the analysis. As our induction hypothesis, suppose on the event $\mc{Y}$, the claim in Eq.~\eqref{eqn:ind_claim} holds for all $k \in [t].$ To argue that it also holds for iteration $t+1$, we invoke the error decomposition in Lemma~\ref{lemma:errordec} to obtain:
\begin{equation}
\begin{aligned}
         d_{t+1} &\le (1-\alpha)^{t+1}d_0 + \gamma \sum_{k=0}^{t} \alpha (1-\alpha)^{t-k} d_k + A_{1,t} + A_{2,t}\\
         & \le (1-\alpha)^{t+1} d_0 + \underbrace{\alpha \gamma \sum_{k=0}^{t} (1-\alpha)^{t-k}(1-\alpha(1-\gamma))^{k} d_0}_{(**)}\\
              & \hspace{3mm} + \alpha \frac{\gamma W}{1-\gamma} \sum_{k=0}^{t} (1-\alpha)^{t-k}  + W\\
        & \le (1-\alpha)^{t+1}d_0 + (**) + \alpha \frac{\gamma W}{1-\gamma} \sum_{k=0}^{\infty} (1-\alpha)^{k} + W\\
        & =  (1-\alpha)^{t+1}d_0 + (**) + \frac{\gamma W}{1-\gamma} + W\\
        & = (1-\alpha)^{t+1}d_0 + (**) + \frac{W}{1-\gamma},
\end{aligned}
\label{eqn:ind_intermed1}
\end{equation}
where for the second inequality, we used the induction hypothesis in tandem with the fact that on the event $\mc{Y}$, $A_{1,t}+A_{2,t} \leq W, \forall t \in [T]$. Now observe that
\begin{equation}
\begin{aligned}
(**) &= \alpha \gamma (1-\alpha)^t d_0 \sum_{k=0}^{t} \left(1+ \frac{\alpha \gamma}{(1-\alpha)}\right)^k\\
&=(1-\alpha)^{t+1}d_0 \left[ \left(1+ \frac{\alpha \gamma}{(1-\alpha)}\right)^{t+1}-1           \right]\\
&=(1-\alpha (1-\gamma))^{t+1}d_0 - (1-\alpha)^{t+1}d_0.
\end{aligned}
\end{equation}
Combining the above display with Eq.~\eqref{eqn:ind_intermed1} establishes the induction claim for iteration $t+1$. We have essentially argued that the bound in Eq.~\eqref{eqn:ind_claim} holds for all $t\in [T]$ on the event $\mc{Y}$. The claim we set out to prove follows by noting that $\mc{Y}$ has measure at least $1-\delta$. All that remains now is to simplify the bound in Eq.~\eqref{eqn:ind_claim} with $t=T$ and $\alpha$ set to $\frac{\log T}{(1-\gamma)T}$. Using $1-x \leq \exp(-x), \forall x \geq 0$, we obtain $(1-\alpha(1-\gamma))^T d_0 = d_0/T.$ Furthermore, a bit of simple algebra reveals that
$$\frac{W}{1-\gamma} = O\left( \frac{\R}{(1-\gamma)^{\frac{5}{2}}}  \frac{\log T}{\sqrt{T}} \sqrt{ \log \left(\frac{|\mathcal{S}||\mathcal{A}|T}{\delta}\right)} +  \frac{\R\sqrt{\varepsilon}}{1-\gamma}\right).$$
This completes the proof of Theorem~\ref{theorem:theoremmainresult}. 
\end{proof}

\section{Tackling Unbounded Reward Distributions}
In this section, we will significantly relax the assumption of the rewards being bounded. In fact, we will not even require the true reward distributions to be sub-Gaussian. Let us now formalize the reward model. In the absence of corruptions, as before, an agent gets to observe a noisy reward $r(s,a) \sim \mc{R}(s,a)$ upon playing action $a$ in state $s$. Moreover, $R(s,a) \triangleq \mathbb{E}_{r(s,a) \sim \mc{R}(s,a)}[r(s,a)]$, and $\mathbb{E}_{r(s,a) \sim \mc{R}(s,a)}[(r(s,a)- R(s,a))^2] \leq \sigma^2.$ In words, playing action $a$ in state $s$ generates a noisy reward random variable $r(s,a)$ with mean $R(s,a)$ and variance upper-bounded by $\sigma^2.$ The key departure from the model in Section~\ref{sec:prob_form} is that we no longer require $r(s,a)$ to be bounded; this naturally rules out the possibility of using trivial thresholding algorithms that ignore rewards falling outside a certain interval since the reward distribution $\mc{R}(s,a)$ can potentially have an infinite support. Moreover, since we only require finiteness of the second moment, the reward distributions can be \emph{heavy-tailed}. \emph{This makes it particularly challenging to distinguish between a corrupted reward sample and a clean reward sample drawn from the tail of the distribution.} Nonetheless, with little to no modifications to our developments thus far, we can handle the challenging setting described above. To make this precise, we make the basic assumption that the means of all reward functions are uniformly bounded\footnote{Note that this should not be confused with the noisy reward realizations being uniformly bounded.}; without such an assumption, one cannot even guarantee the finiteness of the state-action value functions. Accordingly, let $B \geq 1$ be such that $|R(s,a)| \leq B, \forall (s,a) \in \mc{S} \times \mc{A}.$ If we re-define $\bar{r}$ as $\max\{B, \sigma\}$, and plug in this definition of $\bar{r}$ into the threshold in Eq.~\eqref{eqn:Gt}, \emph{then with no further modifications to the algorithm or analysis}, our main result in Theorem~\ref{theorem:theoremmainresult} goes through. This follows by simply revisiting the argument in Case II of Lemma~\ref{lemma:adversarialbnd}. \textbf{The main message here is that even in the absence of corruptions, one can employ our proposed algorithm to handle heavy-tailed rewards, i.e., the additional robustness to heavy-tailed rewards comes essentially for free with our approach.}

\label{sec:unboundedrewards}

\section{Conclusion}
We studied model-free synchronous Q-learning under a strong-contamination reward attack model, and developed a robust algorithm that provides near-optimal guarantees. There are several immediate directions that are part of our ongoing work: establishing fundamental lower bounds, considering the effect of asynchronous sampling, and studying the function approximation setting. In addition, the design of our algorithm requires knowledge of an upper bound on the means and variances of the reward distributions. It would be interesting to see if one can continue to achieve the bounds in this paper without such knowledge; this appears to be quite non-trivial. 
\bibliographystyle{unsrt}
%\setcitestyle{authoryear,open={square} and close={square}}
\bibliography{refs}
\end{document}